\documentclass[runningheads]{llncs}

\usepackage[T1]{fontenc}
\usepackage{graphicx}
\usepackage{amsmath}
\usepackage{amssymb}
\usepackage{url}
\usepackage{xcolor}
\usepackage{marvosym}
\definecolor{linknavy}{rgb}{0,0,0.55}
\usepackage[colorlinks=true,linkcolor=linknavy,citecolor=linknavy,urlcolor=linknavy]{hyperref}
\begin{document}

% GLED-Bench preprint (signed) — built from the DASFAA 2027 anonymous submission
% （中段 usepackage 区原样不动）

\title{GLED-Bench: A Unified Benchmark for Graph-Level Anomalous Event Detection on Dynamic Graphs}

\authorrunning{S. Yang et al.}
\titlerunning{GLED-Bench: A Benchmark for Graph-Level Anomalous Event Detection}
%
% ===== 署名版（预印）=====
\author{Suchang Yang\,\inst{1} \and Hongtao Yu\,\Letter\inst{1} \and
        Ruiyang Huang\inst{1} \and Ran Li\inst{1} \and Junzheng Li\inst{1}}
\institute{Information Engineering University, Zhengzhou, China\\
\email{yuhongtao202508@163.com}}

\maketitle

\begin{abstract}
Graph-level anomalous event detection on dynamic graphs is the task of deciding, at each time step of a discrete-time dynamic graph (DTDG), whether the whole snapshot deviates from the routine dynamics of the sequence. The task serves applications from cybersecurity to disaster response, yet it lacks a dedicated benchmark. Existing temporal graph benchmarks target link and node property prediction, dynamic graph anomaly detection benchmarks score individual nodes or edges, and graph-level anomaly detection benchmarks remain static or rely on injected anomalies, so natively labeled event data and a common protocol for this task are missing. We present GLED-Bench, to our knowledge the first step-wise benchmark for the task on DTDGs with native per-step labels. It assembles seven real-world event datasets from four domains, a statistical rule that tags each dataset as stationary, drifting, or periodic by its event process, a causal evaluation protocol with temporal splits and validation-only model selection, and twelve baseline methods in four categories. The evaluation shows that no method or temporal-modeling choice dominates, as the best-performing method changes from dataset to dataset and the best-performing method category differs across regimes on the sequences studied. GLED-Bench provides a common ground for evaluating detectors of graph-level anomalous events.
\end{abstract}


\section{Introduction}\label{sec:intro}

Graph-level anomalous event detection on dynamic graphs identifies the time steps at which system-level anomalous events occur, and supports cybersecurity, disaster response \cite{dyged}, social unrest monitoring \cite{icews}, and urban mobility \cite{nyctlc}. Dynamic graphs divide into two forms, continuous-time dynamic graphs (CTDGs) based on streaming interactions and discrete-time dynamic graphs (DTDGs) partitioned by time steps. By its step-wise nature, graph-level event detection falls on the DTDG form. At each time step, the detector decides whether the graph of that step, taken as a whole, is an anomalous event that deviates from the routine dynamics of the sequence, rather than scoring individual nodes or edges \cite{dyged}. Unlike node- and edge-level anomaly detection \cite{Hua2026BAGBA}, graph-level detection captures what per-entity scores cannot: it detects system-wide disruptions that surface only as coordinated shifts in the aggregate interaction structure, with every node and edge remaining within its normal range. Graph-level detection on DTDGs therefore requires its own models and evaluation protocols.

However, benchmarks for dynamic graph-level anomalous event detection are still lacking. Existing temporal graph benchmarks \cite{tgb,tgb2,dgb,tgrab,tgbseq} are built around link prediction and node property prediction, and research on dynamic graph anomaly detection, together with its dedicated benchmarks, largely focuses on node- and edge-level detection \cite{zheng2019addgraph,tian2023sad,10.1145/3637528.3671845,generaldyg,Ekle_2024,Hua2026BAGBA}; both lines score individual entities rather than whole snapshots, so neither serves a task whose decision unit is the graph at one time step. Graph anomaly detection beyond the dynamic setting, whether at node level \cite{gadbench} or graph level \cite{glockd,ubgold}, has concentrated on static graphs, where each graph is judged against other graphs rather than against the routine dynamics of its own sequence, and recent dynamic benchmarks that do cover graph-level granularity still evaluate anomaly patterns injected into normal sequences \cite{dygadbench,maritime} rather than events recorded from real systems; natively labeled event sequences, where each step carries a label derived from real event records, remain scarce. Moreover, the few studies that address graph-level event detection directly evaluate their own dataset selections under incomparable splits, features, and metrics \cite{dyged,dygcl}, and a recent audit shows that even in static graph-level detection, evaluation choices such as model selection with test labels can inflate reported scores by a wide margin \cite{glad_reeval}. Graph-level anomalous event detection thus remains without a benchmark, natively labeled data, or a common protocol.

To fill this gap, we present GLED-Bench, a unified benchmark for graph-level anomalous event detection on DTDGs. GLED-Bench encompasses 7 real-world event sequences with native per-step labels, drawn from 4 domains and spanning 3 dynamical regimes (stationary, drifting, and periodic), together with a standardized evaluation protocol that uses causal temporal splits, fixed seeds, and graph-level metrics without point adjustment. On GLED-Bench, we benchmark 12 representative methods covering the shallow, static graph, graph-level dynamic, and RNN-based categories.

Our contributions are summarized as follows:
\begin{itemize}
\item We propose GLED-Bench, to our knowledge the first step-wise benchmark for graph-level anomalous event detection on DTDGs with native per-step labels, assembling 7 real-world sequences from 4 domains whose event processes span 3 regimes (stationary, drifting, and periodic).
\item We design a causal evaluation protocol that removes the common sources of incomparability in prior studies, with strictly temporal splits, unified feature conventions, fixed seeds, and graph-level metrics computed without point adjustment.
\item We conduct a systematic evaluation of 12 representative methods from four categories under a common input convention and protocol, and provide a regime-level analysis that characterizes when method categories succeed or fail across stationary, drifting, and periodic sequences.
\end{itemize}

\section{Preliminaries}\label{sec:prelim}

\subsection{Discrete-Time Dynamic Graphs}\label{sec:prelim:dtdg}

A discrete-time dynamic graph (DTDG) is a sequence of graph snapshots $\mathcal{G} = (G_1, G_2, \ldots, G_T)$ observed on a fixed time grid, where each snapshot $G_t = (V_t, E_t, X_t)$ records the active nodes $V_t$, the interactions $E_t$ among them, and node attributes $X_t$ at time step $t$.

\subsection{Graph-Level Anomalous Event Detection}\label{sec:prelim:task}

Given a DTDG $\mathcal{G}$, the detector outputs a score $s_t$ (equivalently, a binary decision) for every time step $t$, indicating whether the graph of that step, taken as a whole, is an anomalous event; the decision unit is the entire snapshot $G_t$ rather than any individual node or edge. Every time step of the sequence carries a binary label $y_t \in \{0,1\}$ obtained from external event records; training uses the labeled steps of the training segment, and detection is evaluated on temporally later steps.

An anomalous event is a time step whose snapshot is driven by an external event and deviates from the routine dynamics of the sequence, that is, from the behavior the system exhibits in steps not marked by such events. The label $y_t = 1$ marks the steps whose window coincides with an external event record, such as a red-team compromise, a weather warning, a protest episode, or a stadium event. We make no assumption that anomalous steps are rare; their rate varies widely across sequences.

\section{Related Work}\label{sec:related}

\subsection{Anomaly Detection on Dynamic Graphs}\label{sec:related:dygad}

Anomaly detection on dynamic graphs has been studied mainly at the node and edge levels. NetWalk updates node embeddings incrementally and scores vertices through dynamic clustering \cite{netwalk}. AddGraph attaches an attention-based temporal GCN to spot anomalous edges \cite{zheng2019addgraph}. TADDY encodes structural and temporal node roles with a dynamic graph transformer and scores nodes and edges jointly \cite{taddy}. SAD considers the semi-supervised variant of the task \cite{tian2023sad}. SLADE learns self-supervised edge representations for unlabeled interaction streams \cite{10.1145/3637528.3671845}, and GeneralDyG samples temporal ego-graphs to train detectors that transfer across datasets \cite{generaldyg}. A survey consolidates this method landscape \cite{Ekle_2024}. Recent benchmarks organize it: BAG evaluates 25 methods at the node and edge levels on 10 datasets and reports CTDG encoders to be stronger than DTDG encoders for edge anomalies \cite{Hua2026BAGBA}; DyGADBench contributes a method taxonomy and evaluates detection against injected anomaly patterns \cite{dygadbench}; a maritime benchmark spans node-, edge-, and graph-level granularities with LLM-synthesized anomalies \cite{maritime}. Across these methods and benchmarks, the scored unit is a node or an edge, and graph-level granularity appears only with injected anomalies.

\subsection{Graph-Level Anomaly Detection}\label{sec:related:glad}

Graph-level anomaly detection is an established field on static graph collections. GLocalKD distills global and local normal patterns jointly and scores a graph against the other graphs of its dataset \cite{glockd}. OCGTL formulates the task as one-class classification over learned graph transformations \cite{ocgtl}. ANEMONE contrasts node-scale and graph-scale views to detect both locally and globally anomalous graphs \cite{anemone}. GADBench benchmarks supervised detection at the node level \cite{gadbench}, and UB-GOLD unifies graph-level anomaly and out-of-distribution detection over 35 static datasets \cite{ubgold}. An audit of released code shows that selecting checkpoints with test labels raises the mean AUROC of 12 detectors on UB-GOLD by more than five points \cite{glad_reeval}. Graph-level treatment of dynamic data is rare: GraphAnoGAN ranks snapshots of an attributed graph stream with a generative adversarial framework, without event labels \cite{graphanogan}. Throughout this line, a graph is judged against a population of graphs rather than against the routine dynamics of a sequence.

\subsection{Event Detection on Dynamic Graphs}\label{sec:related:event}

Event studies on dynamic graphs originate in societal event forecasting. Deng et al. learn dynamic context graphs from ICEWS-derived word co-occurrence networks to forecast protest events \cite{dygcnn}. DyGED turns to detection: it correlates a sequence of graph-level representations with labeled events and evaluates on two sequences, taxi trips and weather mentions \cite{dyged}. DyGCL stays with forecasting, combining local and global view encoders under contrastive learning over six datasets \cite{dygcl}. Forecasting predicts future labeled steps, whereas detection scores each observed step; the two have not been evaluated under a common protocol.

\section{GLED-Bench}\label{sec:bench}

This section presents the architecture, datasets, baselines, and evaluation protocol of GLED-Bench.

\subsection{GLED-Bench Architecture}\label{sec:bench:arch}

\begin{figure}[t]
\centering
\includegraphics[width=0.92\textwidth]{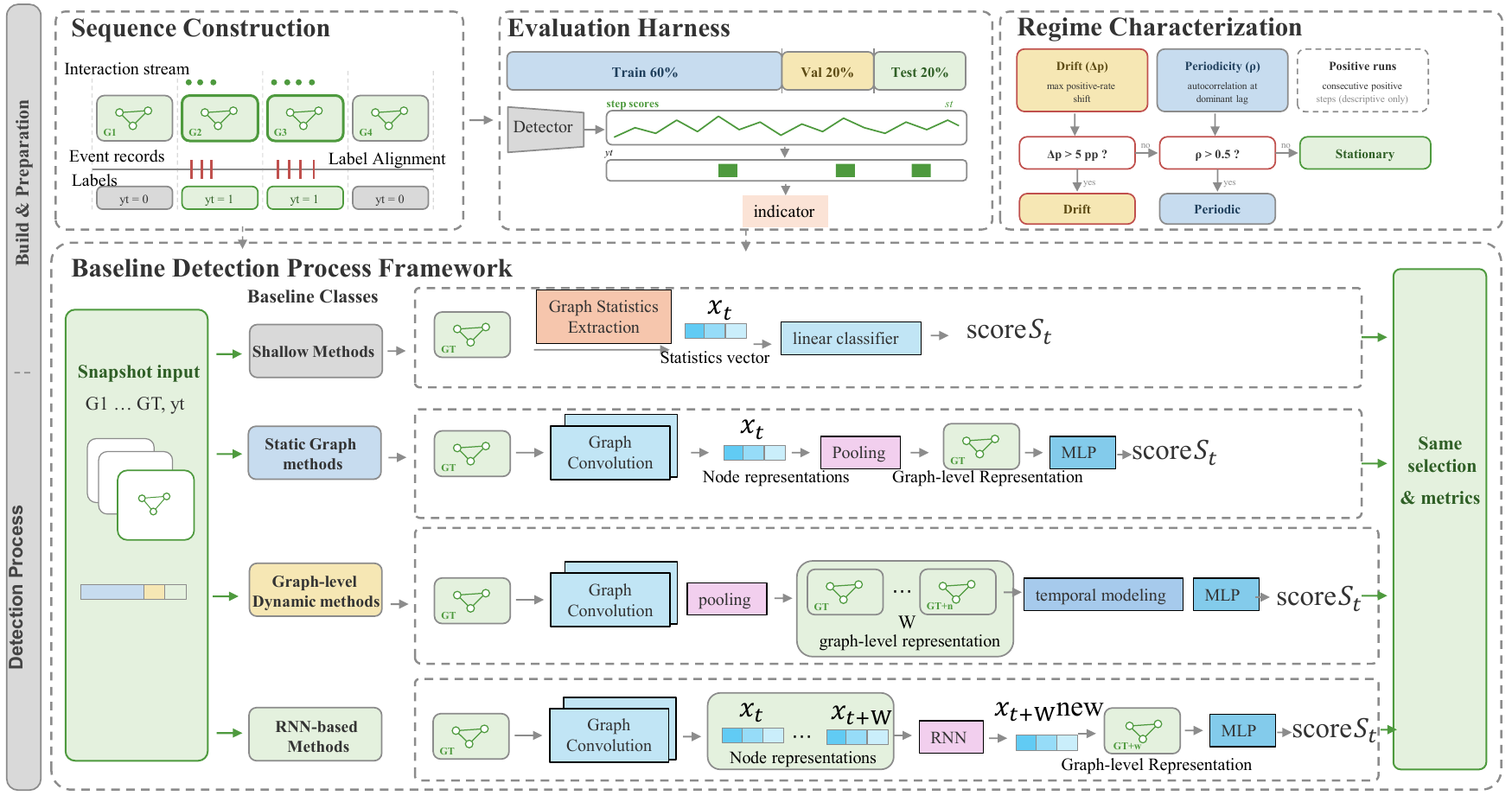}
\caption{Architecture of GLED-Bench.}
\label{fig:overview}
\end{figure}

The architecture of GLED-Bench is shown in Fig.~\ref{fig:overview}. Given a raw event stream and its external event records, we first align the stream on a fixed time grid and aggregate the interactions into graph snapshots. Next, we align the event records to the same grid, which assigns each step a binary label and yields the labeled DTDG of Sect.~\ref{sec:prelim:dtdg}. We then feed the labeled sequence to the evaluation harness, which splits it into segments, trains a method under test, and scores every remaining step. Finally, we characterize the event process with three statistics and tag its dynamical regime. The four components are described in turn.

\paragraph{Sequence Construction.} The raw interactions arrive as a stream $\mathcal{S} = \{(u_i, v_i, \tau_i)\}_{i=1}^{N}$ of triplets with continuous timestamps. With grid length $\Delta$, step $t$ covers the interval $I_t = ((t-1)\Delta,\, t\Delta]$, and the snapshot aggregates the interactions of its cell into weighted edges,
\begin{equation}
w_t(u,v) = \left| \{\, i : (u_i, v_i) = (u, v),\ \tau_i \in I_t \,\} \right|, \qquad E_t = \{\, (u,v) : w_t(u,v) > 0 \,\}.
\end{equation}
The node set $V_t$ either collects the active nodes of the step or stays fixed across the sequence, and the attributes $X_t$ follow the original release of the source, augmented with dynamic structural features for the methods that require them.

\paragraph{Label Alignment.} The external event records form a set $\mathcal{R}$ of timestamps on their own clock. We align them to the same grid, and the step label is
\begin{equation}
y_t = \mathbf{1}\left[\, \exists\, \tau \in \mathcal{R} : \tau \in I_t \,\right],
\end{equation}
where $\mathbf{1}[\cdot]$ is 1 when its argument holds and 0 otherwise, so a step is positive when its window overlaps at least one record. All labels are native, and no anomalies are injected into any sequence.

\paragraph{Evaluation Harness.} The harness partitions $\{1, \ldots, T\}$ into contiguous training, validation, and test segments $S_{\mathrm{train}}, S_{\mathrm{val}}, S_{\mathrm{test}}$ in temporal order. For a target step $t$, it assembles the sample window $(G_{t-W+1}, \ldots, G_t)$ with label $y_t$, where $W$ is the window length of the method under test, and the method returns the step score $s_t = \phi(G_{t-W+1}, \ldots, G_t)$, where $\phi$ denotes the detector. The detector trains on the labeled steps of the training segment and scores every step of the validation and test segments, and the harness stores the results per seed.

\paragraph{Regime Characterization.} To evaluate the baselines under distinct dynamical conditions, we group the sequences into three regimes, stationary, drifting, and periodic, according to the structure of their event process. We ask two questions about the events: how they surface in the graph, and when they occur, and answer both with twelve per-step graph statistics, covering edge count, volume, node count, density, degree and edge-weight quantiles, and edge-turnover rates; no score uses the test segment, and the exact twelve statistics are listed in the released artifact.

The drift score $d$ measures the stability of the way events surface. For a statistic $f$, let $\mathrm{AUC}(f, S)$ be its AUC against the labels on segment $S$, with sign $+1$ if the AUC is at least $1/2$, and let $\mathrm{dir}_{\mathrm{train}}(f)$ be the majority of its sign over four subsegments of the training segment. Then
\begin{equation}
d = \frac{\left|\{\, f \in \mathcal{U} : \mathrm{sign}\,\mathrm{AUC}(f, S_{\mathrm{val}}) \neq \mathrm{dir}_{\mathrm{train}}(f) \,\}\right|}{|\mathcal{U}|},
\end{equation}
the fraction of statistics whose validation direction contradicts the training majority, where $\mathcal{U}$ collects the statistics evaluable on both segments. The training segment is split into four equal contiguous subsegments, a tied majority direction counts as positive, and a statistic is evaluable when its validation segment holds steps of both classes. A high $d$ marks sequences where the way events surface flips across segments.

The periodicity score $\pi$ and the isolation score $\iota$ measure when the events occur:
\begin{equation}
\pi = \max_{f \in \mathcal{F}} \left[ \gamma_f(P) - \gamma_f\!\left(\tfrac{P}{2}\right) \right], \qquad \iota = \frac{\left|\{\, t \in Y^{+} : y_{t-1} = 0 \,\wedge\, y_{t+1} = 0 \,\}\right|}{|Y^{+}|},
\end{equation}
where $\gamma_f(\ell)$ is the autocorrelation at lag $\ell$ of $f$ after subtracting a moving average of window $2P+1$, computed on the nonempty steps of the training and validation segments, $\mathcal{F}$ collects the twelve statistics, and $Y^{+}$ collects the positive steps of the training segment. $P$ is the dominant cycle lag of the grid, the daily lag of 24 steps on the hourly grids and 4 on the six-hourly grids, and the weekly lag of 7 on the daily grid. A high $\pi$ marks a cyclic background, and a high $\iota$ marks events that occur as isolated single steps; periodicity requires both, since a cyclic background alone does not imply that the events recur on the cycle.

A sequence is drifting when the way events surface is unstable across segments, periodic when isolated single-step events recur on a cycle, and stationary otherwise:
\begin{equation}
\mathrm{tag} = \begin{cases} \text{drifting}, & d > \tfrac{1}{2}, \\ \text{periodic}, & \pi > \tfrac{1}{2} \,\wedge\, \iota > 0.8, \\ \text{stationary}, & \text{otherwise}. \end{cases}
\end{equation}
The three thresholds carry meaning rather than being fitted. The drift threshold $\tfrac{1}{2}$ is a majority rule: fewer than half flipped statistics count as noise around a stable relation, and only a majority flip marks the relation as unstable. The periodicity threshold $\tfrac{1}{2}$ sits far outside the $\pm 1.96/\sqrt{n}$ fluctuation band that white noise produces at these sequence lengths, so only a substantial cycle passes it. The isolation threshold $0.8$ is a purity requirement: periodic means that the events are almost all isolated single steps, and a substantial share of runs would break that description.

\subsection{Datasets}\label{sec:bench:data}

GLED-Bench contains seven sequences from four domains, cybersecurity, disaster response, social unrest, and urban mobility.

\begin{itemize}
\item \textbf{LANL} records the authentication stream of an enterprise network from the LANL comprehensive cyber-security release \cite{kent2015lanl}. The grid is one hour. A directed edge $(u, v)$ counts the authentication events from computer $u$ to computer $v$ within the hour, and the step label marks whether at least one red-team compromise event falls inside it.
\item \textbf{NYC Cab} contains taxi trips among the 263 zones of New York City, released with DyGED \cite{dyged}. The grid is one hour. An edge counts the trips from one zone to another within the hour, and a step is positive when a stadium game takes place \cite{nyctlc}.
\item \textbf{Twitter Weather} contains co-occurrence networks over 300 weather words, also released with DyGED \cite{dyged}. The grid is one day. An edge counts the co-occurrence of two words on the day, and a step is positive when the National Weather Service reports a weather event.
\item \textbf{Thailand, Egypt, Russia, and India} contain word co-occurrence networks built from ICEWS news text for Thailand, Egypt, Russia, and India, released with DynamicGCN \cite{dygcnn,icews}. The grid is six hours. Each snapshot unions the per-layer graphs of the release with the maximum weight, and a step is positive when a protest-type golden event falls inside the window.
\end{itemize}

\begin{table}[t]
\caption{Statistics of the GLED-Bench datasets.}
\label{tab:datasets}
\centering\scriptsize
\begin{tabular}{lcccc}
\hline
Dataset & Domain & $T$ & Nodes & Pos.\ rate \\
\hline
LANL & cybersecurity & 711 & 9{,}024 & 11.8\% \\
NYC Cab & urban mobility & 4{,}464 & 263 & 3.6\% \\
Twitter Weather & disaster response & 2{,}557 & 300 & 11.2\% \\
Thailand & social unrest & 1{,}883 & 532 & 37.6\% \\
Egypt & social unrest & 3{,}788 & 282 & 39.3\% \\
Russia & social unrest & 3{,}552 & 390 & 33.2\% \\
India & social unrest & 12{,}249 & 513 & 37.1\% \\
\hline
\end{tabular}
\end{table}

Table~\ref{tab:datasets} reports the statistics of the seven sequences. They span 711 to 12{,}249 steps, their positive rates range from 3.6\% to 39.3\%, and their node sets are fixed in NYC Cab and Twitter Weather and vary per step in the other five. LANL is truncated to its first 711 steps (about 30 days), which contain all red-team event steps of the release; the remaining 681 steps carry no positives and would leave the validation and test segments unscorable. Node counts are averaged over steps for the variable-node sequences.

\subsection{Baselines}\label{sec:bench:baselines}

GLED-Bench targets the supervised setting, where labeled training steps are available; unsupervised and one-class detectors that judge each snapshot against a normal population operate under a different assumption set and are out of scope. The baseline pool contains twelve methods in four categories, and all of them run through the same harness on the same inputs. The categories cover non-neural references, static graph classification models, dynamic methods that build a graph-level representation of each snapshot, and dynamic methods that couple graph encoders with recurrent networks.

\paragraph{Shallow methods.} These methods apply simple classifiers to handcrafted graph statistics and serve as non-neural references. Size-LR and Delta-LR feed size- and change-based statistics to logistic regression, and NetSimile-XGB feeds NetSimile-style descriptors to gradient-boosted trees \cite{netsimile,xgboost}.

\paragraph{Static graph methods.} These methods apply graph classification models to each snapshot in isolation, so they see the structure of one step but no temporal context. GCN \cite{kipf2017} encodes the snapshot with graph convolutions and a readout layer, and TopKPool \cite{gao2019} and SAGPool \cite{lee2019} replace the readout with hierarchical pooling before classification.

\paragraph{Graph-level dynamic methods.} These methods encode each snapshot into a graph-level representation and model the sequence of such representations. DynamicGCN \cite{dygcnn} evolves node embeddings with a temporal layer and reads them out per snapshot, DyGED \cite{dyged} applies structural and temporal self-attention to the graph macro dynamics, and DyGCL \cite{dygcl} combines a local node-level encoder with a global top-k pooling view under contrastive learning.

\paragraph{RNN-based methods.} These methods couple graph encoders with recurrent networks, a standard design for discrete-time dynamic graphs. T-GCN \cite{tgcn} combines graph convolutions with a GRU, GCN-LSTM, the LSTM counterpart studied in \cite{tgcn}, combines them with an LSTM, and EvolveGCN \cite{evolvegcn} evolves the graph convolution weights with a GRU. EvolveGCN applies to the two sequences released with fixed node sets, following the applicability stated in its original release.

Each method follows the architecture and training recipe of its official release, adapted to the common harness and the step-wise scoring interface of Sect.~\ref{sec:bench:protocol}, and the per-method adaptations are documented in the released artifact.

\subsection{Evaluation Protocol}\label{sec:bench:protocol}

This subsection fixes the evaluation rules and the configuration space of GLED-Bench for every baseline on every sequence.

\paragraph{Protocol.} Each sequence is split into training, validation, and test segments in temporal order with a 60/20/20 ratio, and steps with empty snapshots are excluded from evaluation. Every configuration runs with three random seeds, and LANL uses the truncated release of Sect.~\ref{sec:bench:data}. The shallow methods operate on the statistics of the current step, the static graph methods receive a single snapshot per sample, the graph-level dynamic methods receive a window whose length is selected as described in Sect.~\ref{sec:exp:sensitivity}, and the RNN-based methods receive the two-step window of their released configurations; searching their window lengths on the same grid is left to future work. The node axis follows DyGCL, with a full node set when the sequence has at most 1{,}000 nodes and the active nodes of each step otherwise. Training is capped at 100 epochs and stops early when the validation AUC does not improve for 20 epochs, and the reported model is the validation-best epoch of the selected configuration. Detection is scored on the test segment with AUC and average precision (AP). We do not apply point adjustment, since it relabels whole runs of steps around a single hit and would credit methods for steps they missed.

\paragraph{Hyperparameters.} To stay consistent with the baselines, the architecture hyperparameters of every method follow its official implementation. The grid searches the remaining hyperparameters, including the window length of the window-based methods; their ranges and values are listed in Table~\ref{tab:grid}. The configuration of each method on each dataset is selected by the mean validation AUC over the three seeds, and test labels are never used for selection, an integrity concern raised by recent audits of graph-level evaluation \cite{glad_reeval}.

\begin{table}[t]
\caption{Hyperparameter search ranges.}
\label{tab:grid}
\centering\small
\resizebox{\textwidth}{!}{%
\setlength{\tabcolsep}{4pt}
\begin{tabular}{lccc}
\hline
Method & Searched & Fixed (official) & Configs \\
\hline
Size-LR / Delta-LR & $C \in \{0.01, 0.1, 1, 10\}$; class weight $\in$ \{none, balanced\} & -- & 8 \\
NetSimile-XGB & depth $\in \{3, 6\}$; trees $\in \{100, 300\}$; pos.\ weight $\in \{1, \text{auto}\}$ & learning rate 0.1 & 8 \\
GCN & lr $\in \{10^{-2}, 5{\times}10^{-3}, 10^{-3}, 10^{-4}\}$ & hidden 64, dropout 0.5 & 4 \\
TopKPool & lr $\in \{10^{-2}, 5{\times}10^{-3}, 10^{-3}, 10^{-4}\}$ & dropout 0.3 & 4 \\
SAGPool & lr $\in \{10^{-2}, 5{\times}10^{-3}, 10^{-3}, 10^{-4}\}$ & hidden 64, ratio 0.5, dropout 0.5 & 4 \\
DyGED & lr $\in \{10^{-2}, 5{\times}10^{-3}, 10^{-3}, 10^{-4}\}$; $W \in \{1,\ldots,7\}$ & dropout 0.1 & 28 \\
DyGCL & hidden $\in \{32, 64\}$; ratio $\in \{0.5, 0.25\}$; lr $\in \{10^{-2}, 5{\times}10^{-3}, 10^{-3}, 10^{-4}\}$; $W \in \{1,\ldots,7\}$ & dropout 0.2 & 112 \\
DynamicGCN & lr $\in \{10^{-2}, 5{\times}10^{-3}, 10^{-3}, 10^{-4}\}$; $W \in \{1,\ldots,7\}$ & 7 layers, dropout 0.2 & 28 \\
T-GCN & hidden $\in \{8, 16, 32, 64, 100, 128\}$; lr $\in \{10^{-2}, 5{\times}10^{-3}, 10^{-3}, 10^{-4}\}$ & dropout 0; two-step window; weight decay $1.5\times10^{-3}$ & 24 \\
GCN-LSTM & hidden $\in \{64, 128\}$; lr $\in \{10^{-2}, 5{\times}10^{-3}, 10^{-3}, 10^{-4}\}$ & dropout 0; two-step window & 8 \\
EvolveGCN & hidden $\in \{32, 64, 128\}$; lr $\in \{10^{-2}, 5{\times}10^{-3}, 10^{-3}, 10^{-4}\}$ & dropout 0; two-step window & 12 \\
\hline
\end{tabular}}
\end{table}

\section{Experiments}\label{sec:exp}

In this section, we tag the seven sequences into regimes, evaluate the twelve baselines under the protocol of Sect.~\ref{sec:bench:protocol}, and read the results by regime, window sensitivity, and cost.

\subsection{Regime Characterization}\label{sec:exp:regime}

We apply the route of Sect.~\ref{sec:bench:arch} to the seven datasets. As shown in Table~\ref{tab:regime}, on the drift score, Twitter Weather stands far above all others and passes the majority threshold, while every other sequence stays near zero; its feature directions flip across segments, so it is the only drifting sequence. On the periodicity score, only NYC Cab and LANL pass the threshold by a wide margin, while the other five stay near zero, so these two alone carry a substantial cycle. On the isolation score, NYC Cab again stands far above the rest; its events are almost all isolated single steps, and isolated events recurring on a cycle make it the periodic sequence. LANL falls far below the purity threshold, its events arrive in runs, so it is stationary together with the four ICEWS datasets, whose scores stay near zero throughout; its event rate also fluctuates across segments, but the way its events surface stays stable. The tag is computed on the training and validation segments and enters no training, selection, or scoring decision.

\begin{table}[t]
\caption{Routing statistics of the seven datasets, following Sect.~\ref{sec:bench:arch}.}
\label{tab:regime}
\centering
\scriptsize
\begin{tabular}{lcccc}
\hline
Dataset & $d$ & $\pi$ & $\iota$ & Tag \\
\hline
LANL & 0.00 & 1.45 & 0.15 & stationary \\
NYC Cab & 0.00 & 1.43 & 0.96 & periodic \\
Twitter Weather & 0.83 & $-$0.02 & 0.16 & drifting \\
Thailand & 0.00 & $-$0.01 & 0.39 & stationary \\
Egypt & 0.08 & $-$0.04 & 0.21 & stationary \\
Russia & 0.00 & $-$0.06 & 0.43 & stationary \\
India & 0.00 & $-$0.01 & 0.38 & stationary \\
\hline
\end{tabular}
\end{table}

\subsection{Overall Results}\label{sec:exp:overall}

We compare the twelve baselines on test AUC and AP in Tables~\ref{tab:main-auc} and~\ref{tab:main-ap}. Rows are grouped by method category. The window length is selected on validation data: DyGED uses $W=2$ on NYC Cab, DyGCL uses $W=7$ on LANL and Twitter Weather, and all other cells of the three window-based methods use $W=1$.

\begin{table}[t]
\caption{Test AUC (mean$\pm$std over three seeds) of the twelve baselines. The best method per dataset is in bold and the second best is underlined.}
\label{tab:main-auc}
\centering\footnotesize
\resizebox{\textwidth}{!}{%
\setlength{\tabcolsep}{3pt}
\begin{tabular}{lccccccc}
\hline
Method & LANL & NYC Cab & Twitter W. & Thailand & Egypt & Russia & India \\
Size-LR & \underline{0.946\,$\pm$\,0.000} & 0.886\,$\pm$\,0.000 & \underline{0.740\,$\pm$\,0.000} & 0.801\,$\pm$\,0.000 & 0.818\,$\pm$\,0.000 & 0.819\,$\pm$\,0.000 & 0.667\,$\pm$\,0.000 \\
Delta-LR & 0.860\,$\pm$\,0.000 & 0.872\,$\pm$\,0.000 & 0.711\,$\pm$\,0.000 & 0.728\,$\pm$\,0.000 & 0.710\,$\pm$\,0.000 & 0.787\,$\pm$\,0.000 & 0.646\,$\pm$\,0.000 \\
NetSimile-XGB & 0.916\,$\pm$\,0.000 & \underline{0.899\,$\pm$\,0.000} & 0.540\,$\pm$\,0.000 & 0.791\,$\pm$\,0.000 & 0.846\,$\pm$\,0.000 & 0.812\,$\pm$\,0.000 & 0.682\,$\pm$\,0.000 \\
\hline
GCN & \textbf{0.951\,$\pm$\,0.004} & 0.863\,$\pm$\,0.004 & \textbf{0.761\,$\pm$\,0.059} & \textbf{0.843\,$\pm$\,0.012} & 0.912\,$\pm$\,0.004 & \underline{0.882\,$\pm$\,0.003} & 0.726\,$\pm$\,0.003 \\
TopKPool & 0.941\,$\pm$\,0.010 & 0.795\,$\pm$\,0.005 & 0.666\,$\pm$\,0.058 & 0.837\,$\pm$\,0.023 & 0.906\,$\pm$\,0.005 & 0.873\,$\pm$\,0.005 & 0.730\,$\pm$\,0.003 \\
SAGPool & 0.918\,$\pm$\,0.017 & 0.745\,$\pm$\,0.028 & 0.654\,$\pm$\,0.097 & 0.837\,$\pm$\,0.011 & \underline{0.914\,$\pm$\,0.003} & 0.871\,$\pm$\,0.012 & 0.747\,$\pm$\,0.004 \\
\hline
DynamicGCN & 0.920\,$\pm$\,0.004 & 0.809\,$\pm$\,0.028 & 0.704\,$\pm$\,0.038 & 0.775\,$\pm$\,0.028 & 0.773\,$\pm$\,0.038 & 0.809\,$\pm$\,0.015 & 0.663\,$\pm$\,0.006 \\
DyGED & 0.934\,$\pm$\,0.019 & \textbf{0.932\,$\pm$\,0.004} & 0.643\,$\pm$\,0.086 & 0.805\,$\pm$\,0.029 & 0.872\,$\pm$\,0.001 & 0.858\,$\pm$\,0.013 & 0.726\,$\pm$\,0.004 \\
DyGCL & 0.880\,$\pm$\,0.011 & 0.845\,$\pm$\,0.003 & 0.686\,$\pm$\,0.066 & 0.830\,$\pm$\,0.017 & \textbf{0.916\,$\pm$\,0.003} & 0.879\,$\pm$\,0.009 & 0.739\,$\pm$\,0.002 \\
\hline
T-GCN & 0.907\,$\pm$\,0.028 & 0.841\,$\pm$\,0.009 & 0.703\,$\pm$\,0.014 & 0.828\,$\pm$\,0.003 & 0.907\,$\pm$\,0.001 & 0.876\,$\pm$\,0.004 & \textbf{0.753\,$\pm$\,0.009} \\
GCN-LSTM & 0.924\,$\pm$\,0.005 & 0.698\,$\pm$\,0.032 & 0.528\,$\pm$\,0.031 & \underline{0.842\,$\pm$\,0.010} & 0.911\,$\pm$\,0.002 & \textbf{0.893\,$\pm$\,0.004} & \underline{0.752\,$\pm$\,0.012} \\
EvolveGCN & -- & 0.870\,$\pm$\,0.030 & 0.268\,$\pm$\,0.044 & -- & -- & -- & -- \\
\hline\end{tabular}}
\end{table}

\begin{table}[t]
\caption{Test AP (mean$\pm$std over three seeds) of the twelve baselines, under the same layout as Table~\ref{tab:main-auc}.}
\label{tab:main-ap}
\centering\footnotesize
\resizebox{\textwidth}{!}{%
\setlength{\tabcolsep}{3pt}
\begin{tabular}{lccccccc}
\hline
Method & LANL & NYC Cab & Twitter W. & Thailand & Egypt & Russia & India \\
Size-LR & 0.603\,$\pm$\,0.000 & 0.222\,$\pm$\,0.000 & \textbf{0.443\,$\pm$\,0.000} & 0.745\,$\pm$\,0.000 & 0.855\,$\pm$\,0.000 & 0.680\,$\pm$\,0.000 & 0.593\,$\pm$\,0.000 \\
Delta-LR & 0.285\,$\pm$\,0.000 & 0.311\,$\pm$\,0.000 & 0.243\,$\pm$\,0.000 & 0.656\,$\pm$\,0.000 & 0.715\,$\pm$\,0.000 & 0.652\,$\pm$\,0.000 & 0.539\,$\pm$\,0.000 \\
NetSimile-XGB & 0.436\,$\pm$\,0.000 & \underline{0.347\,$\pm$\,0.000} & 0.140\,$\pm$\,0.000 & 0.737\,$\pm$\,0.000 & 0.879\,$\pm$\,0.000 & 0.668\,$\pm$\,0.000 & 0.598\,$\pm$\,0.000 \\
\hline
GCN & \textbf{0.635\,$\pm$\,0.014} & 0.157\,$\pm$\,0.016 & \underline{0.394\,$\pm$\,0.044} & 0.793\,$\pm$\,0.005 & 0.939\,$\pm$\,0.004 & 0.809\,$\pm$\,0.010 & 0.635\,$\pm$\,0.008 \\
TopKPool & \underline{0.616\,$\pm$\,0.059} & 0.088\,$\pm$\,0.002 & 0.326\,$\pm$\,0.031 & 0.790\,$\pm$\,0.032 & 0.935\,$\pm$\,0.004 & 0.760\,$\pm$\,0.021 & 0.653\,$\pm$\,0.001 \\
SAGPool & 0.478\,$\pm$\,0.137 & 0.112\,$\pm$\,0.043 & 0.294\,$\pm$\,0.100 & \underline{0.802\,$\pm$\,0.009} & 0.940\,$\pm$\,0.005 & 0.801\,$\pm$\,0.028 & 0.675\,$\pm$\,0.005 \\
\hline
DynamicGCN & 0.471\,$\pm$\,0.075 & 0.127\,$\pm$\,0.040 & 0.249\,$\pm$\,0.043 & 0.686\,$\pm$\,0.103 & 0.786\,$\pm$\,0.054 & 0.662\,$\pm$\,0.058 & 0.561\,$\pm$\,0.015 \\
DyGED & 0.574\,$\pm$\,0.074 & \textbf{0.367\,$\pm$\,0.029} & 0.259\,$\pm$\,0.104 & 0.750\,$\pm$\,0.040 & 0.905\,$\pm$\,0.004 & 0.759\,$\pm$\,0.016 & 0.647\,$\pm$\,0.005 \\
DyGCL & 0.351\,$\pm$\,0.033 & 0.130\,$\pm$\,0.004 & 0.267\,$\pm$\,0.113 & 0.797\,$\pm$\,0.016 & \textbf{0.943\,$\pm$\,0.003} & 0.787\,$\pm$\,0.010 & \underline{0.677\,$\pm$\,0.006} \\
\hline
T-GCN & 0.418\,$\pm$\,0.096 & 0.128\,$\pm$\,0.013 & 0.370\,$\pm$\,0.026 & 0.792\,$\pm$\,0.007 & 0.939\,$\pm$\,0.000 & \underline{0.818\,$\pm$\,0.007} & 0.676\,$\pm$\,0.013 \\
GCN-LSTM & 0.498\,$\pm$\,0.032 & 0.067\,$\pm$\,0.017 & 0.158\,$\pm$\,0.018 & \textbf{0.819\,$\pm$\,0.005} & \underline{0.942\,$\pm$\,0.002} & \textbf{0.835\,$\pm$\,0.020} & \textbf{0.686\,$\pm$\,0.006} \\
EvolveGCN & -- & 0.240\,$\pm$\,0.086 & 0.089\,$\pm$\,0.004 & -- & -- & -- & -- \\
\hline\end{tabular}}
\end{table}

Table~\ref{tab:main-auc} and Table~\ref{tab:main-ap} summarize the overall comparison. No method dominates, as five different methods place first on each metric and the two metrics agree on the best method on four of the seven datasets. GCN is the most consistent method, staying within the top six on test AUC on every dataset, Delta-LR and DynamicGCN occupy the bottom two positions on both metrics on the four ICEWS datasets, and detection difficulty itself varies widely across the datasets, as the best method on LANL achieves the highest AUC of the whole benchmark, while the best methods on India and Twitter Weather achieve the lowest.

Each category shows a distinct strength and a matching failure mode. The shallow methods are cheap and deterministic and perform best wherever events surface in global statistics, but NetSimile-XGB breaks down on Twitter Weather, encoding the sequence drift rather than the events (Sect.~\ref{sec:exp:regimeperf}), and the category falls behind on the hardest stationary dataset, where the event signal is weak in global summaries. The overall best performance of the static graph category comes from plain GCN, which transfers across every regime, while its pooling variants remain competitive on the stationary sequences yet decline on the drifting and periodic ones, consistent with node selection tuned to one stable structure being misled as the snapshot pattern shifts. The graph-level dynamic category spans the widest internal range, from the best performance of DyGCL to the worst of DynamicGCN, as its members differ in how directly they serve the task. The RNN-based category turns temporal context into an advantage on the stationary sequences but loses it on the drifting and periodic ones, where the carried state plausibly tracks the training segment, so GCN-LSTM performs far worse and EvolveGCN, restricted to the fixed-node datasets, fails outright with inverted score directions (Sect.~\ref{sec:exp:regimeperf}). On several datasets the top methods are not separated beyond seed variability, so the per-dataset winners should be read as pool-level observations rather than separable rankings.

\subsection{Performance by Regime}\label{sec:exp:regimeperf}

We aggregate test AUC by method category and regime (Table~\ref{tab:regime-perf}); the best-performing category differs across the three regimes.

\begin{table}[t]
\caption{Mean test AUC by method category and regime, averaged over the applicable method-dataset cells of Table~\ref{tab:main-auc}. For the stationary column, $\pm$ gives the standard deviation across the five datasets; the drifting and periodic columns hold a single dataset each, and their seed-level variability is reported per cell in Tables~\ref{tab:main-auc} and~\ref{tab:main-ap}. The best category per regime is in bold, ties included; the stationary tie between the static and RNN categories is within display precision. EvolveGCN is excluded from the RNN-based aggregates.}
\label{tab:regime-perf}
\centering\footnotesize
\begin{tabular}{lccc}
\hline
Category & Stationary & Drifting & Periodic \\
\hline
Shallow & 0.789\,$\pm$\,0.09 & 0.664 & \textbf{0.886} \\
Static graph & \textbf{0.859\,$\pm$\,0.08} & \textbf{0.694} & 0.801 \\
Graph-level dynamic & 0.825\,$\pm$\,0.08 & 0.678 & 0.862 \\
RNN-based & \textbf{0.859\,$\pm$\,0.07} & 0.616 & 0.770 \\
\hline
\end{tabular}
\end{table}
\begin{table}[t]
\caption{Score-level diagnostics on Twitter Weather: $\rho_t$ is the Spearman correlation between the test score and time, and the last column recomputes the test AUC of Table~\ref{tab:main-auc} after removing the linear time trend.}
\label{tab:diag}
\centering\scriptsize
\begin{tabular}{lcc}
\hline
Method & $\rho_t$ & AUC after detrending \\
\hline
Size-LR & 0.45 & 0.685 \\
Delta-LR & 0.37 & 0.696 \\
NetSimile-XGB & 0.24 & 0.502 \\
GCN & 0.20 & 0.730 \\
TopKPool & $-$0.03 & 0.680 \\
SAGPool & 0.13 & 0.633 \\
DynamicGCN & 0.22 & 0.650 \\
DyGED & $-$0.01 & 0.652 \\
DyGCL & 0.22 & 0.649 \\
T-GCN & 0.10 & 0.680 \\
GCN-LSTM & $-$0.09 & 0.541 \\
EvolveGCN & 0.18 & 0.225 \\
\hline
\end{tabular}
\end{table}
Following the regimes of Sect.~\ref{sec:exp:regime}, each group of datasets shows a distinct pattern. On the five stationary sequences the static graph and RNN-based categories tie for the best performance and the graph-level dynamic category follows closely, so where the event process carries no strong temporal signature the choice among them matters little. NYC Cab is the periodic sequence and the only dataset where a graph-level dynamic method performs best, DyGED with a two-step window, though the category aggregate of Table~\ref{tab:regime-perf} still puts the shallow category first on this sequence; its events are isolated single steps, so the informative signal is the change from the previous step rather than the level of the current one. Twitter Weather is the drifting sequence with the widest spread of method performance in the benchmark; the static graph methods perform best on it, and the score-level diagnostics below attribute the spread to how each method handles the drift.

The Twitter Weather scores expose three distinct failure patterns (Table~\ref{tab:diag}). Removing the linear time trend from the scores costs every healthy method little of its discrimination, and GCN, the strongest detector here, retains the most. NetSimile-XGB instead falls to chance, so its descriptors encode the drifting word distribution and no event signal, whereas GCN-LSTM carries little drift and no signal at all. EvolveGCN finally scores below chance in every seed, the signature of an inverted score direction rather than a missing one.

\subsection{Sensitivity Analysis}\label{sec:exp:sensitivity}

\begin{figure}[t]
\centering
\includegraphics[width=0.92\textwidth]{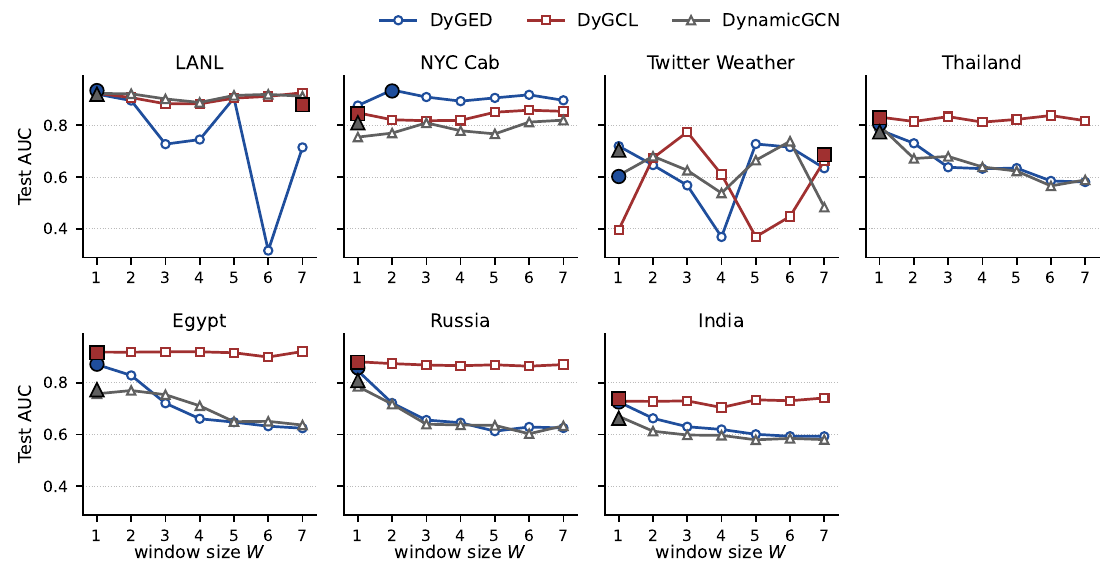}
\caption{Single-seed test AUC against the window size $W$ for the three window-based methods. Marked points indicate the window used in Tables~\ref{tab:main-auc} and~\ref{tab:main-ap}, plotted at their three-seed means.}
\label{fig:wsens}
\end{figure}

We analyze the window length $W$, the key hyperparameter of the three window-based methods, which determines how much history a model observes when scoring a step. Fig.~\ref{fig:wsens} plots single-seed test AUC against $W$ on the seven datasets, and the trends differ across datasets. On the four ICEWS datasets, the AUC of DyGED and DynamicGCN declines steadily as the window grows, and $W=1$ gives the highest validation and test AUC for both, whereas DyGCL is insensitive to the window length. A longer history therefore adds no information beyond the current snapshot on these sequences. On LANL, none of the curves follows a trend, and DyGED becomes unstable at long windows. NYC Cab and Twitter Weather behave differently. On NYC Cab, $W=1$ is the weakest setting of DyGED, and a two-step window improves it markedly. On Twitter Weather, every curve fluctuates with the window length and validation and test disagree, but the three-seed validation check confirms no improvement over the default for DyGED and for DynamicGCN, so both keep the default window.

\subsection{Efficiency}\label{sec:exp:efficiency}

We measure the training cost of each method. Table~\ref{tab:efficiency} reports the training time of the selected configuration from the first epoch until early stopping, averaged over seeds, all measured on NVIDIA RTX A6000 GPUs with Intel Xeon Gold 5218R CPUs. The shallow methods train within seconds. Among the neural methods, cost depends on the method rather than on the category, and it spans nearly three orders of magnitude. DyGED is the cheapest neural method on four datasets, since it reaches its best epoch early and stops soon after, whereas the pooling methods of the static category are the most expensive ones on the four ICEWS sequences and on LANL, with TopKPool taking up to 7.13 hours on Russia. Cost and accuracy do not align: DyGED, the cheapest neural method on four datasets, performs best on NYC Cab, while the most expensive methods rank in the middle.

\begin{table}[t]
\caption{Total training time in hours of the selected configuration until early stopping, averaged over three seeds. Shallow methods train within seconds ($<$0.01~h); EvolveGCN applies to the fixed-node datasets only.}
\label{tab:efficiency}
\centering\scriptsize
\resizebox{\textwidth}{!}{%
\setlength{\tabcolsep}{3pt}
\begin{tabular}{lccccccc}
\hline
Method & LANL & NYC Cab & Twitter W. & Thailand & Egypt & Russia & India \\
GCN & 1.35 & 0.01 & $<$0.01 & 0.02 & 2.04 & 1.62 & 2.37 \\
TopKPool & 2.40 & 0.37 & 0.12 & 0.31 & 2.85 & 7.13 & 4.74 \\
SAGPool & 3.71 & 0.02 & $<$0.01 & 0.02 & 2.59 & 2.46 & 1.60 \\
\hline
DynamicGCN & 0.10 & 0.92 & 0.35 & 0.25 & 0.36 & 3.43 & 3.31 \\
DyGED & 0.02 & 0.33 & $<$0.01 & 0.02 & 0.08 & 0.03 & 0.16 \\
DyGCL & 0.27 & 0.59 & 0.51 & 0.10 & 0.13 & 0.11 & 0.24 \\
\hline
T-GCN & 0.25 & 0.50 & 0.30 & 0.14 & 0.15 & 0.50 & 0.41 \\
GCN-LSTM & 0.12 & 1.86 & 0.51 & 0.27 & 0.94 & 1.85 & 0.97 \\
EvolveGCN & -- & 0.68 & 0.16 & -- & -- & -- & -- \\
\hline\end{tabular}}
\end{table}

\section{Conclusion}\label{sec:concl}

We presented GLED-Bench, a unified benchmark for graph-level anomalous event detection on discrete-time dynamic graphs, built from seven natively labeled event datasets across four domains and three dynamical regimes, with a causal evaluation protocol and twelve baseline methods in four categories. The evaluation shows that no method or temporal-modeling choice dominates the task, as the best-performing method changes from dataset to dataset and the best-performing category differs across regimes on the sequences studied. The regime statistics that tag the datasets involve no detector outcome, so the grouping can be reused on labeled sequences, and the reversal of the category ranking on the drifting sequence shows that a ranking obtained on one sequence can reverse on another. The scope of GLED-Bench is supervised step-wise detection on DTDGs, and extending the dataset pool and the method pool within the same protocol is the natural next step.

\bibliographystyle{splncs04}
\bibliography{mybibliography}

\end{document}